\documentclass[times,referee,twocolumn,final,authoryear]{elsarticle}

\usepackage{ycviu}
\usepackage{framed,multirow}

\usepackage{amssymb}
\usepackage{latexsym}

\usepackage{url}
\usepackage{xcolor}
\definecolor{newcolor}{rgb}{.8,.349,.1}

\newcommand{\etal}{\textit{et al}.}

\usepackage{subcaption}
\usepackage[ruled,vlined,noend]{algorithm2e}
\include{pythonlisting}
\usepackage{times}
\usepackage{epsfig}
\usepackage{graphicx}
\usepackage{amsmath}
\usepackage{amssymb}
\usepackage{mathabx}
\usepackage{epsfig}
\usepackage{times}
\usepackage{multirow}
\usepackage{hhline}
\usepackage{graphics}
\usepackage{comment}
\usepackage{anyfontsize}
\usepackage{url}

\usepackage{rotating}
\usepackage{float}
\usepackage{listings}
\usepackage{courier}
\usepackage{multicol}
\usepackage{hyperref}

\SetKwInput{KwInput}{Input}
\SetKwInput{KwPre}{Pre-train}

\usepackage{algorithmic}
\newlength\myindent
\newcommand\norm[1]{\left\lVert#1\right\rVert}

\newenvironment{indentblock}{%
  \par%
  \leftskip=1.25em%
  \noindent}{%
  \par}

\journal{Computer Vision and Image Understanding}

\begin{document}

\ifpreprint
  \setcounter{page}{1}
\else
  \setcounter{page}{1}
\fi

\begin{frontmatter}

\title{Grow-Push-Prune: aligning deep discriminants for effective structural network compression}

\author[1]{Qing \snm{Tian}\corref{cor1}}
\cortext[cor1]{Corresponding author: 
  }
\ead{qing.tian@mail.mcgill.ca}
\author[1]{Tal \snm{Arbel}}
\author[1]{James J. \snm{Clark}}

\address[1]{Centre for Intelligent Machines, McGill University, 3480 University Street, Montreal, QC H3A 0E9, Canada}

\received{1 May 2013}
\finalform{10 May 2013}
\accepted{13 May 2013}
\availableonline{15 May 2013}
\communicated{S. Sarkar}

\begin{abstract}
Most of today's popular deep architectures are hand-engineered to be generalists. However, this design procedure usually leads to massive redundant, useless, or even harmful features for specific tasks. Unnecessarily high complexities render deep nets impractical for many real-world applications, especially those without powerful GPU support. In this paper, we attempt to derive task-dependent compact models from a deep discriminant analysis perspective. We propose an iterative and proactive approach for classification tasks which alternates between (1) a pushing step, with an objective to simultaneously maximize class separation, penalize co-variances, and push deep discriminants into alignment with a compact set of neurons, and (2) a pruning step, which discards less useful or even interfering neurons. Deconvolution is adopted to reverse `unimportant' filters' effects and recover useful contributing sources. A simple network growing strategy based on the basic Inception module is proposed for challenging tasks requiring larger capacity than what the base net can offer. Experiments on the MNIST, CIFAR10, and ImageNet datasets demonstrate our approach's efficacy. On ImageNet, by pushing and pruning our grown Inception-88 model, we achieve more accurate models than Inception nets generated during growing, residual nets, and popular compact nets at similar sizes. We also show that our grown Inception nets (without hard-coded dimension alignment) clearly outperform residual nets of similar complexities.
\end{abstract}

\begin{keyword}
\MSC 41A05\sep 41A10\sep 65D05\sep 65D17
\KWD deep neural network pruning \sep deep discriminant analysis\sep deep representation learning

\end{keyword}

\end{frontmatter}


\section{Introduction}

Compact yet capable neural network architectures are desirable for many real-world problems, such as HCI, autonomous driving perception, and video analytics.
Many network pruning approaches proposed so far pay little attention to whether the complexity decrease follows a task-optimal direction, such as those based on weight magnitudes \citep{han20150}. Moreover, most of them are ex post facto, i.e., useful and useless components are already mixed and it is too late to trim one without influencing the other.
Aside from pruning, a compact structure design practice is to utilize a random number of $1\times1$ filters, usually at module ends to reduce feature map dimension \citep{he2015,szegedy2015,iandola2016,howard2017}. Nevertheless, an ad-hoc filter number may lead to irrecoverable information loss or redundancy/overfitting/interference.

In this paper, we propose to derive task-suitable compact networks through deep discriminant analysis in the feature space. Instead of counting on an optimally pre-trained model, the proposed approach follows a two-step procedure in iterations. (1) through learning, it proactively unravels useful twisted threads of deep variation and pushes them into alignment with a compact substructure that can be easily decoupled from the rest.
(2) with important features being held separated from the rest, the second pruning step simply throws away the inactive, useless, or even harmful features over the layers. Cross-layer dependency is tracked by deconvolution based utility reconstruction. We push and prune in a progressive and gradual manner since it helps improve and expedite the convergence at each iteration. We will show, through solving a generalized eigenvalue problem, that the first step can be achieved by simultaneously including deep LDA and covariance penalty terms to the optimization objective. The LDA and covariance loss terms are calculated per batch at the easily disentangled end (final latent space), but exert influence over the layers. For scenarios where the desired capacity is larger than what the base structure can offer, a simple network growing/expansion strategy is proposed.
In contrast to fixed network architectures, our grow-push-prune pipeline provides an approach capable of generating a range of task-suitable models for different needs and constraints. 

It is worth mentioning that this work is fundamentally different from our previous work \citep{tian2021}. \cite{tian2021} is a passive after-the-fact pruning approach that targets a pre-trained model. There is no alignment of discriminants with easily pruned structures. \cite{tian2021} cannot prune much if useful and useless features are already intertwined in the base. On the other hand, the proposed framework here proactively maximizes, condenses, and separates useful information flows over the network that contribute to the final class separation. Also, we propose a useful base net growing strategy while \cite{tian2021} can do nothing if the base net capacity is not large enough (one-way top-down search).

In our experiments on the MNIST, CIFAR10, and ImageNet datasets, efficient compact models with comparable/better accuracies to the base can be derived. In the ImageNet case, our series of grown deep Inception nets beat residual structures at similar complexities without any hard-coded dimension alignment. One of our grown deep Inception net, Inception-88, beats ResNet-50 (slightly larger) after training with the conventional cross-entropy and $L_2$ losses. Deep LDA pushing not only pushes utility into alignment with a compact set of latent neuron dimensions but also further increases the accuracy by 0.2\%. The pruning step based on the `pushed' model leads to a series of compact models with accuracies even higher than our grown deep Inception nets and similar-sized resnets. At a pruning rate of approximately 6\%, a pruned model achieves accuracy 0.4\% higher than the original unpruned Inception-88.

\section{Related Work}

\paragraph{Neural Networks Pruning}

Early pruning approaches targeted at shallow nets date back to the late 1980s~\citep{pratt1989,lecun1989prune,hassibi1993,reed1993}. \cite{reed1993} offers a review for such early researches targeted at shallow nets.
Aimed at deep nets, \cite{han20150} abandon weights of small magnitude by setting them to zero.
Similar approaches that sparsify networks by setting zeros include~\citep{srinivas2015,mariet2016,jin2016,guo2016,hu2016,sze2017}.
\cite{frankle2019} hypothesize that a large neural network contains a smaller subnetwork (winning ticket) which, when trained separately, can achieve similar accuracy. The top-down manner of search is necessary.
Weights based pruning usually leads to unstructured sparsity. 
More recently, filter/neuron/channel pruning has gained popularity (e.g.~\cite{polyak2015,anwar2015,li2016pruning,tian2017,he2017channel,zhuang2018,molchanov2019,he2019,he2020learning,chin2020towards,guo2020dmcp,wang2021convolutional}). Instead of setting zeros in weights matrices, they remove rows, columns, depths in weight/convolution matrices. Thus, the resulting architectures are more hardware friendly. They require not only less storage space and transportation bandwidth, but also less computation. Moreover, with fewer intermediate feature maps produced and consumed, the number of slow and energy-consuming memory accesses is decreased.
That said, most pruning works possess some of the following drawbacks: (1) hard-coded utilities are computed locally and not directly related to final classification, such as magnitude and variance of weights and activation. (2) they usually depend on a pre-trained or passively learned model. It may be too late to prune after the fact that useful and harmful components are already intertwined together. (3) some filter-based approaches, such as~\cite{zhuang2018,molchanov2019,chin2020towards}, rely on an implicit weight-level independence assumption (e.g. $L_p$-norms).
In this paper, we propose a deep discriminant analysis based importance measure that takes into consideration the relationships between weights, filters, and across layers. Unlike most existing pruning importance measures, our measure is directly related to the final class separation power. We proactively align the neuron outputs with discriminant directions through training the network before discarding useless, redundant, or interfering dimensions.
Aside from pruning, approaches like bitwise reduction~\citep{rastegari2016,han2015,sun2016,gupta2015,gong2014,courbariaux2015}, filter decomposition~\citep{denton2014exploiting,jaderberg2014speeding,zhang2016accelerating}, knowledge distillation~\citep{hinton2015distilling}, and depth-wise separable convolution~\citep{chollet2017xception,howard2017} can further reduce model complexity. That said, they are beyond the paper's scope.

Growing a network before pruning can sometimes be beneficial~\citep{yuan2020growing,mixter2020growing,huang2005generalized,narasimha2008integrated}. \cite{wang2017growing} demonstrate that more layers can lead to better clustered concepts. \cite{belilovsky2019greedy} show progressive linear separability with the depth increase.

\paragraph{Efficient Neural Architecture Search}

Most of AutoML or Neural Architecture Search (NAS) approaches fall into one of the two categories: reinforcement learning (policy gradient) based~\citep{baker2016,zoph2016neural,zoph2017learning,zhong2017} and evolutionary or genetic algorithms based~\citep{stanley2002,xie2017,miikkulainen2017,real2017large,real2018regularized}. Strict constraints are usually applied to reduce the search space. That said, each sampled architecture still needs to be trained separately. Given the large number of possible architecture samples, the procedure is very computationally expensive. For example, the search processes in \cite{zoph2016neural} and \cite{real2017large,real2018regularized} took the authors 28 days on 800 GPUs and one week on 450 GPUs, respectively. Most such works are done on the small CIFAR10 dataset. When it comes to larger datasets, resulting structures from small datasets are usually stacked up. 
Rather than design the entire network, some start with a macro architecture and fill in different substructure samples into each cell (micro search). ENAS~\citep{pham2018efficient} makes a strong assumption that common structures share the same weights. Similarly, instead of fully training all samples, PNAS~\citep{liu2018progressive} `predicts' the accuracy based on the differences between the new and parent samples.
Bottom-up search in infinite spaces could possibly miss an `optimal' structure in the early stage and never come back to it.
\cite{he2018amc} propose AutoML to search compact models. They trained a reinforcement learning agent to predict layerwise channel shrinking actions. To gain efficiency, the reward is roughly estimated based on the model accuracy prior to finetuning.
Given the large number of architectures sampled, it is no surprise that the best achieves high accuracy.
\cite{liu2019darts} tackle the problem in a differentiable manner and present a weight-sharing method for NAS named DARTS. However, it suffers from performance collapse caused by an inevitable aggregation of skip connections.

\section{Proactive Deep LDA dimension reduction}

Existing AutoML works are generally expensive while passive pruning approaches rely heavily on the base model.
In this paper, we propose a proactive deep discriminant analysis based approach that tracks down task-desirable compact architectures by exploring the deep feature space. Our approach iterates between two steps: (1) maximizing and pushing class separation utility to easily pruned substructures (e.g., neurons) and (2) pruning away less useful substructures. These two steps are illustrated in Algorithm~\ref{alg:proactiveldaprune}, and the details are in Sec.~\ref{subsec:pushingstep} and~\ref{subsec:pruningstep}.

{\centering
\begin{algorithm}
    \small
    \SetAlgoLined
    \KwIn{base model (a popular net or one grown as in Sec.~\ref{sec:compactnetsearch}), acceptable accuracy $t_{acc}$}
    \KwOut{task-suitable compact models}
    \While{True}{
      \vspace{0.05in}
      \textbf{Step 1} $\rightarrow $ \textbf{Pushing}
      \begin{indentblock}
      {
          Train the net with the deep LDA pushing objectives added (red components in Fig.~\ref{fig:LDACovReg});\newline
          \lIf{accuracy $< t_{acc}$}{break}
      }
      \end{indentblock}
      \vspace{0.05in}
      \textbf{Step 2} $\rightarrow $ \textbf{Pruning}
      \begin{indentblock}
      {
          Prune less useful components based on deconv source recovery;
      }
      \end{indentblock}
    }
    \vspace{0.05in}
    \Return{compact models derived}
    \caption{Proactive deep discriminant analysis based pushing and pruning}
\label{alg:proactiveldaprune}
\end{algorithm}
\par
}

\subsection{Pushing step}\label{subsec:pushingstep}

\begin{figure}
  \centering
  \includegraphics[width=0.35\textwidth]{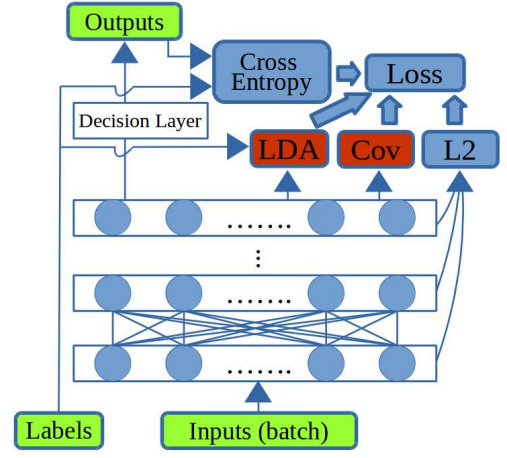}
  \caption{Pushing Step. Our deep LDA push objectives are colored in red. They maximize, unravel, and condense useful information flow transferred over the network and bring discriminants into alignment with latent space neurons. $L_2$ regularization is also applied to the decision layer, but is not shown for clarity.}
\label{fig:LDACovReg}
\end{figure}

The room for complexity reduction in a deep net mainly comes from the useless and redundant structures. Unlike after-the-fact pruning approaches, we explicitly embed these considerations into the loss function. We leverage LDA to boost class separation and utilize covariance losses to penalize redundancies. As we will show later, these terms simultaneously maximize and unravel useful information flow transferred over the network and push discriminant power into a small set of decision-making latent space neurons. The pushing step is demonstrated as Figure~\ref{fig:LDACovReg}.
The LDA and convariance penalty terms are computed at the last latent space (after ReLU) because: (1) it is directly related to decision making and accepts information from all other layers, (2) the linear assumption of LDA is reasonable or easily enforced for over-parameterized networks. 
After all, there is only one linear FC layer left before decision making. Post-decision softmax, if any, is only a monotonic normalization and it cannot change the decision. 
(3) utility can be unraveled with ease from this disentangled or loosely twisted end. That said, these terms, as part of the training objective function, exert influence over the entire network.

Apart from cross-entropy, we explicitly and proactively apply linear discriminant analysis (LDA) in the final latent space to maximize class separation. The goal of the LDA term is to transform data from a noisy and complicated space to one where different categories can be linearly separated (there is only one final FC layer left). It is aligned with the training goal to reduce classification error. Latent features learned are expected to pick up class separating statistics in the input. Inspired by~\cite{fisher1936,rao1948}, we define our deep LDA utility in the final latent space as Eq.~\ref{eq:interintraratiop}, which will take advantage of both deep feature extraction and simple linear separability analysis:

\begin{equation} \label{eq:interintraratiop}
S_{W,\theta} = \frac{\mid W^T\Sigma_{b,\theta}W \mid}{\mid W^T\Sigma_{w,\theta}W \mid}\,,
\end{equation}
where
\begin{equation} \label{eq:intraclassscatter}
\Sigma_{w,\theta} = \sum_{i} \tilde{X}_{\theta,i}^T\tilde{X}_{\theta,i}\,,
\end{equation}
\vspace{-0.5in}
\setlength{\columnsep}{-0.4cm}
\begin{multicols}{2}
  \begin{equation}
    \Sigma_{b,\theta} = \Sigma_{a,\theta} - \Sigma_{w,\theta}\,,
  \end{equation}\break
  \begin{equation}
    \Sigma_{a,\theta} = \tilde{X_\theta}^T\tilde{X_\theta}\,,
  \end{equation}
\end{multicols}
\noindent with $X_{\theta,i}$ being the set of observations obtained in the final latent space for category $i$, with model parameter setting $\theta$. 
A pair of single vertical bars denotes matrix determinant. The tilde sign ( $\tilde{}$ ) represents a centering operation; for data $X$ this means:
\begin{equation}
\tilde{X} = (I_n - n^{-1} 1_n1_n^T) X\,,
\end{equation}
\noindent where $n$ is the number of observations in $X$, $1_n$ denotes an n$\times$1 vector of ones.
The training objective of deep LDA is to maximize the final latent space class separation (Eq.~\ref{eq:interintraratiop}), which comes down to solving the following generalized eigenvalue problem:
\begin{equation} \label{eq:geigenvaluepp}
\Sigma_{b,\theta} \vec{e_j} = v_j \Sigma_{w,\theta} \vec{e_j}\,,
\end{equation}

\noindent where ($\vec{e_j}$,$v_j$) represents a generalized eigenpair of the matrix pencil $(\Sigma_{b,\theta},\Sigma_{w,\theta})$ with $\vec{e_j}$ as a $W$ column. The LDA objective of maximizing Eq.~\ref{eq:interintraratiop} can be achieved by maximizing the average of $v_j$s. Thus, we define the LDA-related loss term as its reciprocal:
\begin{equation} \label{eq:ldaseparation}
\ell_{lda} = \frac{N}{\sum_{j}^N v_j}\,.
\end{equation}
\noindent In practice, we set $N$ as the number of neurons with non-negligible variances (dormant dimensions are not considered). 
Simultaneously, to penalize co-adapted structures and reduce redundancy in the network, we inject covariance penalty into the latent space. The corresponding loss is:
\begin{equation}\label{eq:covterm}
\ell_{cov}  = \norm{\Sigma_{a,\theta} - diag(\Sigma_{a,\theta})}_1\,,
\end{equation}
\noindent where $\norm{.}_1$ indicates entrywise 1-norm. This term agrees with the intuition that, unlike lower layers' common primitive features, higher layers of a well-trained deep net capture a wide variety of high-level, global, and easily disentangled abstractions~\cite{bengio2013,zeiler2014}. Generally speaking, the odds of various high-level patterns firing together should be low.
$\ell_{cov}$ has a side effect of forcing useless dimensions to zero \footnote{For $k$ category classification, there are at most $k-1$ uncorrelated linear discriminants~\citep{hou2015}. $l_{cov}$ reduces redundancy across the dimensions and forces the activations (ideally) only along the ($\leq$) $k-1$ directions.} and thus alleviates over-fitting (similar to dropout, but in a non-random and activation-based way).

Furthermore, to safely prune on the neuron level without much information loss, we need to align the above mentioned LDA utility ($v_j$s) with neuron dimensions. For this purpose, we try to align $W$ columns with standard basis directions, e.g., (1,0,...,0), and let the network learn an optimal $\theta$ that leads to large class separation. This will also save us from using an actual $W$ rotation in addition to the neural net. Given that duplicate neurons have been discouraged by $\ell_{cov}$ and inactive neurons are not considered here, Eq.~\ref{eq:geigenvaluepp} can be rewritten as:
\begin{equation} \label{eq:geigenvaluepplusp}
(\Sigma_{w,\theta}^{-1}\Sigma_{b,\theta}) \vec{e_j} = v_j \vec{e_j}\,.
\end{equation}

\noindent As we can see, $W$ column $\vec{e_j}$s are the eigenvectors of $\Sigma_{w,\theta}^{-1}\Sigma_{b,\theta}$. Thus, forcing the direction alignment of LDA utilities and neuron dimensions is equivalent to forcing $\Sigma_{w,\theta}^{-1}\Sigma_{b,\theta}$ to be a diagonal matrix (eigenvectors of a diagonal matrix form a standard basis). 
We incorporate this constraint by including the following term to the loss function:
\begin{equation}\label{eq:align}
\ell_{align}  = \norm{\Sigma_{w,\theta}^{-1}\Sigma_{b,\theta} - diag(\Sigma_{w,\theta}^{-1}\Sigma_{b,\theta})}_1 \,,
\end{equation}

\noindent where, similar to Eq.~\ref{eq:covterm}, entrywise 1-norm is used instead of entrywise 2-norm (a.k.a. Frobenius norm) because our aim is to put as many off-diagonal elements to zero as possible.
Combining all three terms, we get our pushing objective as follows. Its three components jointly maximize class separation, squeeze and push classification utility into a compact set of neurons for later pruning: 
\begin{equation}\label{eq:pushterms}
\ell_{push} = \gamma\ell_{lda} + \lambda \ell_{cov} + \beta \ell_{align}\,,
\end{equation}
\noindent where $\lambda$, $\beta$, and $\gamma$ are weighting hyperparameters.
Many advanced loss weighting strategies are available (e.g., \cite{kendall2018multi}).
Here, we set them so that (1) LDA utilities and neuron dimensions are aligned and (2) high accuracy is maintained.
In our experiments, through parameter $\theta$ learning in the (large enough) base networks, the two goals can be met simultaneously without much hyperparameter tweaking. With the pushing terms above, we actually obtain higher accuracy on all the datasets explored (Sec.~\ref{sec:proactiveexp}).
In addition to class separation utility boost, the extra constraints can add some structure/regularization to the original overfitted deep space with very high degree of freedom.
These terms help constrain useful information within or near more compact manifolds.
$\ell_{lda}$ is sometimes numerically unstable. Inspired by~\cite{friedman1989}, we add a constant to the diagonal elements of the within-scatter matrix. When the category number is large, we cannot include all categories in one forward pass and the scatter matrices at a certain batch are calculated for a random subset of classes. Each class has the same or similar number of samples ($\geq 8$).
When latent space dimension $d$ is large (e.g., in the first iteration), the $\ell_{align}$ constraint which includes an expensive $d \times d$ matrix inverse operation can be omitted. The reason is that in the context of over-parameterized network and high dimensional latent space, neuron activation is sparse: only a limited number of neurons tend to fire for a class and each high-level neuron motif corresponds to only one or few classes. In this scenario, positive within-class correlation indicates positive total correlation, and minimizing $\ell_{cov}$ has an effect of minimizing $\ell_{align}$. Through training with the pushing objectives added, the network learns to organize itself for easy pruning. $W$ columns that maximize the class separation (Eq.~\ref{eq:interintraratiop}) are aligned with some latent neuron dimensions.
This pushing step lays the foundation for neuron/filter level pruning across all layers.


\subsection{Pruning Step}\label{subsec:pruningstep}

\begin{figure*}[!htp]
\begin{center}
\includegraphics[width=0.8\linewidth]{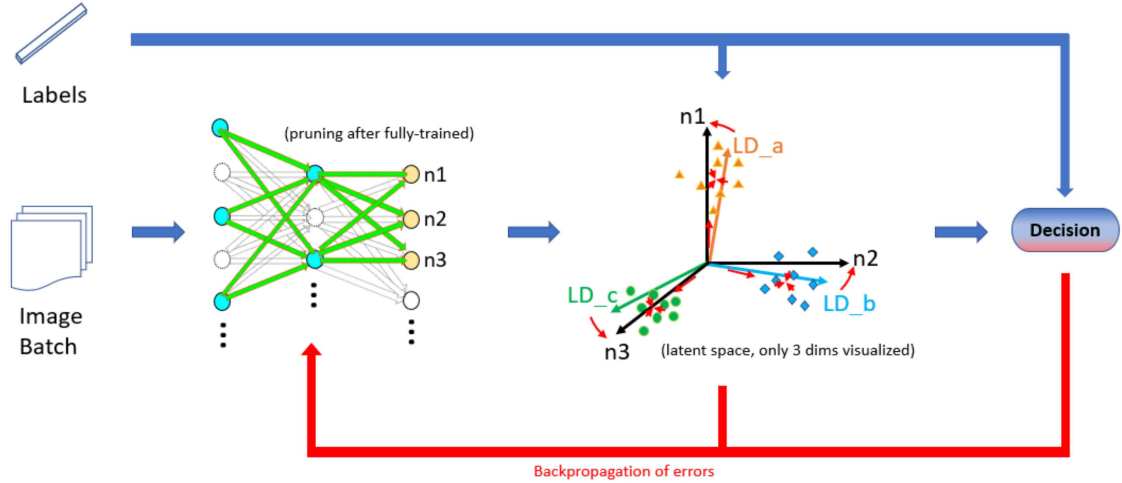}
\end{center}
\vspace{-0.1in}
\caption{Overview of the proposed push-and-prune pipeline. The blue arrows indicate forward information flow while the large red arrows represent backward propagation of errors. In the latent space (only 3 dimensions are visualized), the small red arrows illustrate some effects of our pushing terms (e.g., class separation, neuron-discriminant alignment, and decorrelation of neuron dimensions). Pruning occurs only after a model is fully trained according to our objectives.
White components in the network are of limited utility and can thus be pruned. $n$ indicates final latent space neuron dimensions.
}
\label{fig:overview}
\end{figure*}

We treat pruning as a dimensionality reduction problem in the deep feature space. After the pushing step, the final class separation power is maximized and discriminants are simultaneously pushed into alignment with some top layer neurons. It follows that the direct abandonment of less useful neurons and their dependencies on previous layers is safe. The discriminant power along the $j$th neuron dimension $v_j$ is the corresponding diagonal value of $\Sigma{w}^{-1}\Sigma_{b}$: 
\begin{equation} \label{eq:vjp}
\begin{split}
v_j = diag(\Sigma{w}^{-1}\Sigma_{b})_j \,, 
\end{split}
\end{equation}
\noindent
where $\Sigma{w}$ and $\Sigma_{b}$ are within-class and between-class scatter matrices of the final features based on the parameters trained.
When pruning, we discard neuron dimension $j$s of small $v_j$ along with its cross-layer contributing sources in the `pushed' model where useful components have been separated from others.
To this end, we need to quantitatively measure the utility/contribution of all neurons to final utility before abandoning useless ones. In this paper, we utilize deconvolution/deconv~\citep{zeiler2014,tian2021} to trace the task utility unraveled from final latent space backwards across all layers. In the final layer, only the responses of the most discriminative dimensions are preserved (other dimensions with small $v_j$ are set to 0) before deconv starts.
It is worth mentioning that \cite{zeiler2014} use `deconvolution' for visualization purposes in the image space while we focus on reconstructing contributing sources over the layers. Also, the proposed method only back-propagates useful final variations. Irrelevant and interfering features of various kinds are `filtered out'. The unit deconv procedure performs convolution with the same filters transposed. It can be considered as inversion of convolution with an assumption of orthogonal convolution matrix.
More detail on this part can be found in~\citep{tian2021}.
For modular structures, we need to sum many-to-one dependencies in a module. In other words, the reconstructed utility maps from all module branches need to be summed at the module beginning before the utility can be further traced backwards to previous modules.
With all neurons'/filters' contribution to final discriminability known, pruning simply becomes discarding those of low utility/contribution.
The utility threshold is directly related to the pruning rate. Since feature maps (neuron outputs) correspond to next-layer filter depths (neuron weights), our pruning leads to filter-wise and channel-wise savings simultaneously. After pruning at each iteration, retraining with surviving parameters is needed.
Putting the pieces together, we provide a high-level summary of the proposed push-and-prune pipeline for deep dimension reduction in Figure~\ref{fig:overview}.

\section{Base net growing strategy}\label{sec:compactnetsearch}

One limitation with pruning is that the top-down, one-way search is bounded above by the base net's capacity. For datasets requiring larger capacities than the base net can offer, a growing step before iterative push-and-prune is necessary to first encompass/contain enough sub-architecture candidates. In the language of the famous `lottery ticket hypotheses'~\citep{frankle2019}, the growing step is like `buying more lottery tickets'.
According to \cite{wang2017growing, belilovsky2019greedy}, growing can also lead to better clustered concepts and progressive linear separability.
In our case, it will add extra adjustment/wiggle room for the next push step to disentangle, compress, and re-organize utility in the network.
We grow deep Inception nets by greedily and iteratively adding more modules according to Algorithm~\ref{alg:greedygrow}, which can be viewed as a trial-and-error evolutionary process.

{\centering
\begin{algorithm}
\small
\SetAlgoLined
\DontPrintSemicolon 
\KwIn{$net = \{s_0,s_1,...,s_i,...\}, s_i = \{m_{i0},m_{i1},...,m_{ij},...\}$, where $s$: stage, $m$: module, $net$: starting base. $N$: number of extra modules to add}
\KwOut{net with $N$ extra modules added}
$n=1; acc_{max}=0; net_{opt};$\;
\While{$n \leq N$} {
  \For{$stage$ \textbf{in} $net$}{
    $net'=extend(net, stage)$;\; 
    train $net'$ and predict, get val accuracy $acc$;\;
    \If{$acc > acc_{max}$}{
      $acc_{max} = acc$;\; $net_{opt} = net'$;\;
    }
  }
  $net = net_{opt}$, save if necessary;\; $n = n + 1$;\;
}
\Return{$net$}\;
\caption{Greedy base net growing strategy}
\label{alg:greedygrow}
\end{algorithm}
\par
}
At each iteration, we try to add one module to one of the network stages. Here, a stage consists of several modules with the same output feature map dimension before a pooling layer. 
The newly added module has the same architecture as the module underneath. We quickly train all the possible options (e.g., 3 for the Inception net) and keep only the one that achieves the highest accuracy. The process is repeated for $N$ iterations until reaching a complexity bound or no noticeable accuracy gain can be observed after two consecutive iterations. Like the initial Inception net, when training, two auxiliary classifiers are added to the second stage (one after the first module and the other before the last module). 
We find the auxiliary classifiers very useful when the depth becomes large. 
By this growing strategy, a superset of abundant deep features can be obtained, from which our deep LDA pushing and pruning can derive task-desirable ones.

We prefer the Inception module over ResNets' residual modules because the latter requires hard-coded alignment of dimension, which is known to greatly limit the freedom of pruning~\citep{li2016pruning}.
The skip/residual dimension has to agree with the main trunk dimension for summation. However, after pruning according to any importance measure (including ours), they do not necessarily agree unless we force them to. Given that each ResNet module has only 2-3 layers, such a hard-coded constraint at every module end would greatly limit the freedom of pruning.
Also, compared to residual models, inception modules offer us a variety of filter types. Our deep LDA pruning can take advantage of this by selecting both the numbers and types of filters on different abstraction levels.
Compared to ResNets with up to hundreds of layers, current Inception models are relatively shallow and they only have a dozen or so modules. It has been proven that deep networks are able to approximate the accuracy of shallow networks with an exponentially fewer number of parameters, at least for some classes of functions~\citep{telgarsky2016,eldan2016,mhaskar2016,safran2017,poggio2017}.
In this paper, we explore to grow from the basic inception net~\citep{szegedy2015} by greedily adding more unit modules and see whether the resulting deep Inception nets can achieve ResNet-comparable accuracy.
We use the initial Inception net (a.k.a. GoogLeNet) as the starting point but with two modifications inspired by~\cite{ioffe2015batch}. The first is to approximate the function of $5\times5$ filters with two consecutive $3\times3$ filters, and the second is to add batchnorm after each conv layer. 
In the rest of the paper, when we talk about the Inception module or net, we refer to this variant.
Later inception modules (V2-V4) include more architecture fiddling and usually require higher resolution data (e.g. 299$\times$299). We do not incorporate those changes since we want to perform fair comparisons between our grown deep Inception nets, ResNets, and some other popular networks taking 224 $\times$ 224 input. Also, this keeps human expert knowledge involved as minimum as possible. Ideally, we aim to replace such human knowledge with learning and pruning.
Interestingly, despite its simplicity, no works have explored simply growing the original Inception net to gain accuracy. 
In this paper, we grow the net before deep LDA based pushing and pruning. The results on ImageNet will be shown in Sec.~\ref{sec:proactiveldaimagenet}.

\section{Experiments and results}\label{sec:proactiveexp}

This section tests our proactive deep discriminant analysis based pruning on the MNIST, CIFAR10, and ImageNet datasets. We only perform push and prune on the first two relatively small datasets while apply the entire grow-push-prune pipeline on ImageNet.
For all the datasets, we first train a network (fixed or grown) in the conventional way to report the baseline accuracy before applying the pushing step (with $l_{push}$, defined in Sec.~\ref{subsec:pushingstep}, added to the objective).
There are various non-architectural tricks that can possibly help increase a network's accuracy, such as extra training data, bounding box info, multi-cropping, various input resolutions, label smoothing regularization, mixup training, and distillation~\citep{he2019bag}. 
In our experiments, we refrain from employing such non-architectural tweakings.
Instead of trying to achieve the highest accuracy with all possibly beneficial additions, our focus here is to fairly compare different architectures with few factors of interference.

\subsection{A toy experiment on MNIST}

We use the MNIST dataset to illustrate deep LDA pushing's influence on the latent space. MNIST details can be found in~\cite{lecun1998gradient}. We leave out the first 1,000 images in each category of the training set for validation. With a simple five hidden layer fully-connected network (1024-1024-1024-1024-32), we will show deep LDA pushing's efficacy. In this toy experiment, the last hidden layer is set to have 32 neurons for illustration clarity.

\begin{figure}
\centering
\begin{subfigure}{0.475\linewidth}
    \includegraphics[clip, trim=0 0 0 0,width=\linewidth]{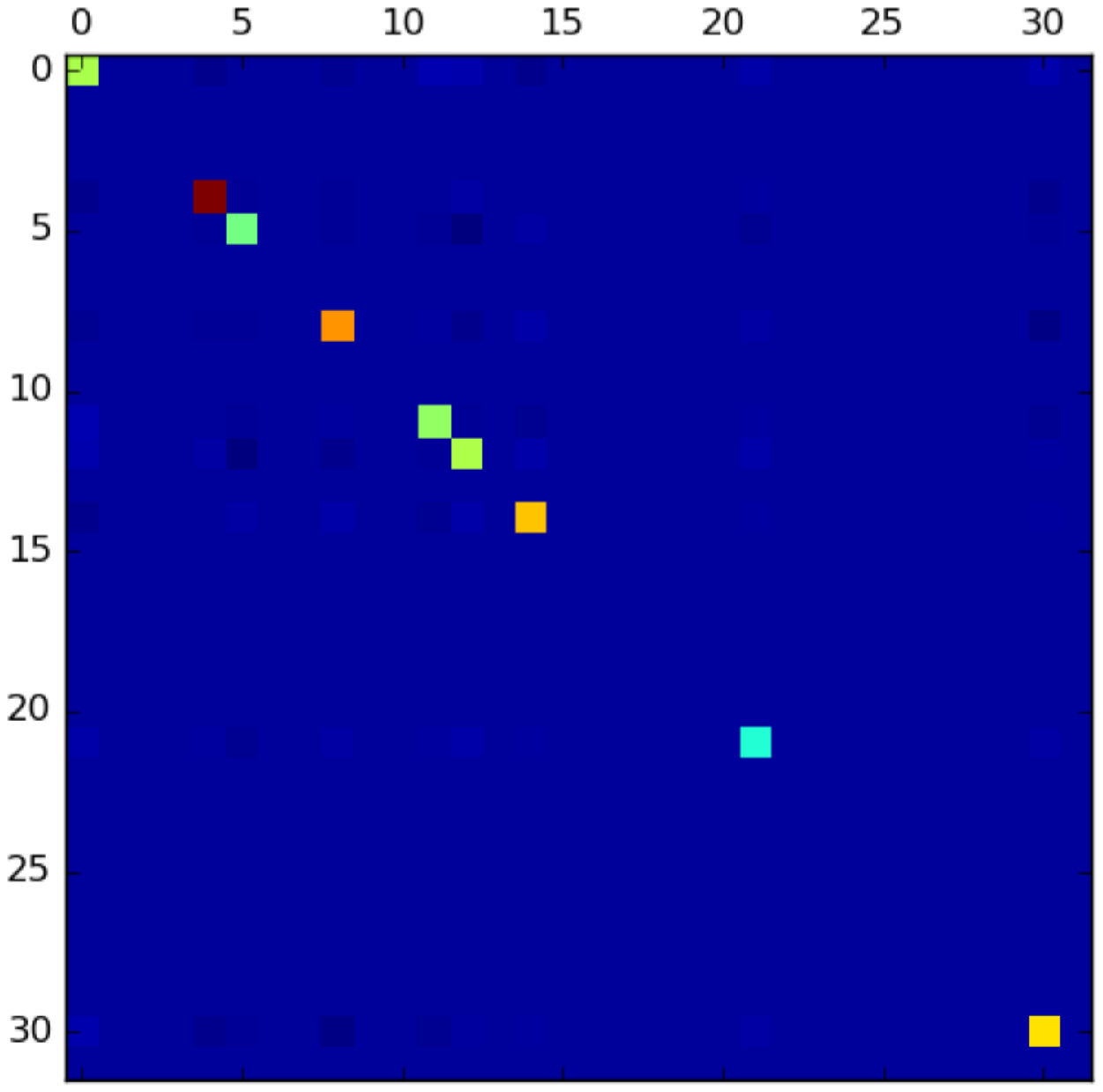}
    \caption{with pushing objective}
    \label{fig:withpush}
\end{subfigure}
~
\begin{subfigure}{0.475\linewidth}
    \includegraphics[clip, trim=0 0 0 0,width=\linewidth]{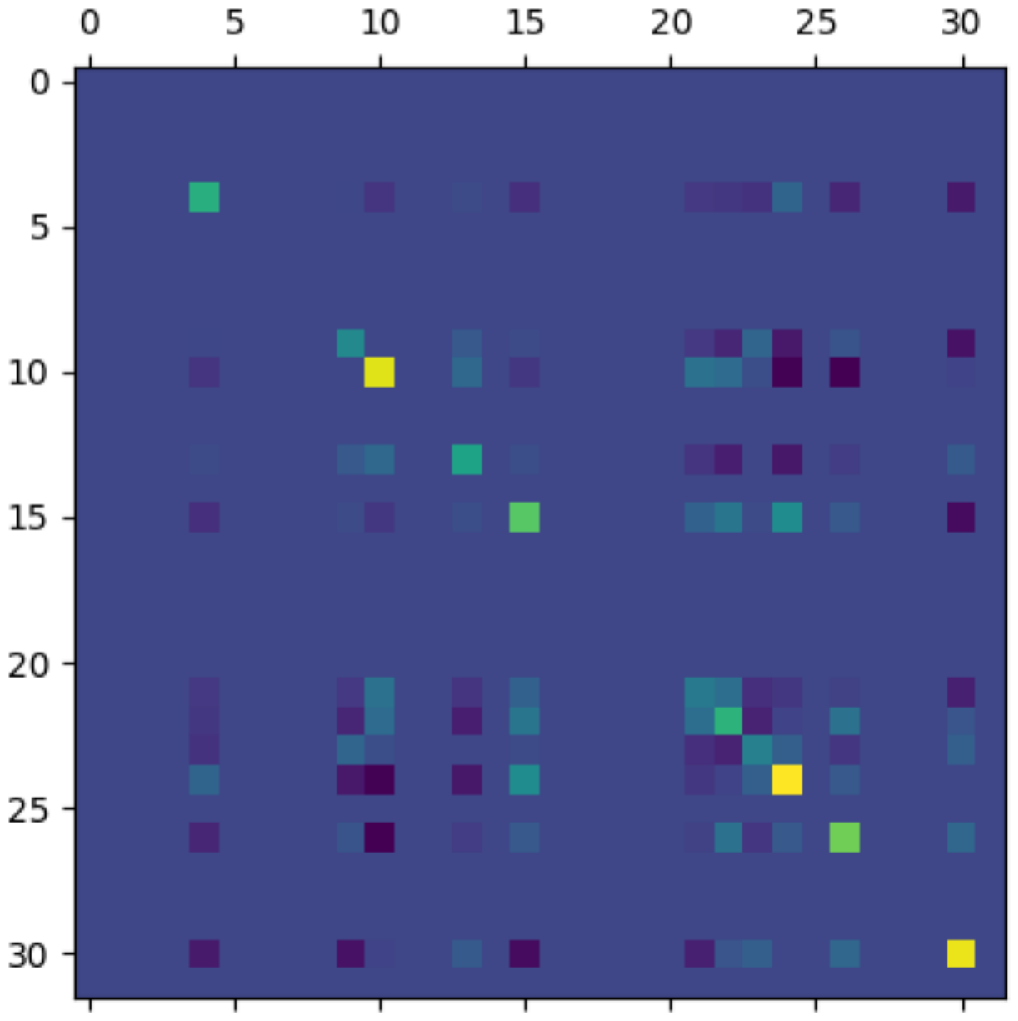}
    \caption{without pushing objective}
    \label{fig:withoutpush}
\end{subfigure}
\caption{Variance-covariance matrices of the latent space neuron output after training (a) with and (b) without the pushing objective (Sec.~\ref{subsec:pushingstep}) on the MNIST dataset using a toy FC architecture (hidden dimensions: 1024-1024-1024-1024-32). The values are color coded using the default bgr color map of the Matplotlib pyplot matshow function~\cite{hunter2007matplotlib}. From small to large, the color transits from blue to green and finally to red.}
\label{fig:pushedcovmnist}
\vspace{-0.1in}
\end{figure}

As mentioned previously, the main purpose of proactive LDA pushing (Sec.~\ref{subsec:pushingstep}) is to push deep discriminants or class separation power into alignment with latent space neuron dimensions so that filter-level pruning is safe. Although the pushing influence is across the layers, here via this toy example, we only illustrate how the final latent space is changed as other layers' changes influence the final decision via this space. Fig.~\ref{fig:pushedcovmnist} visualizes the variance-covariance matrix of latent space neuron output after training with and without the pushing objective.
From Fig.~\ref{fig:pushedcovmnist}, we can see that our proposed deep LDA pushing objective is effective and successfully pushes useful final decision-making variances to a subset of latent dimensions.
Compared to Fig.~\ref{fig:withoutpush}, training with the pushing objective better decorrelates useful variances (Fig.~\ref{fig:withpush}). As aforementioned, this contributes to the alignment of deep discriminants with latent dimensions. Most importantly, the accuracy does not change much by including the pushing objective. In fact, the accuracy even improves slightly. The conventional cross-entropy with $L_2$ leads to a validation accuracy of 97.9\%. This number increases to 98.3\% with the addition of the deep LDA pushing objective.
\begin{figure}
\centering
\begin{subfigure}{0.475\linewidth}
    \includegraphics[clip, trim=0 0 0 0,width=\linewidth]{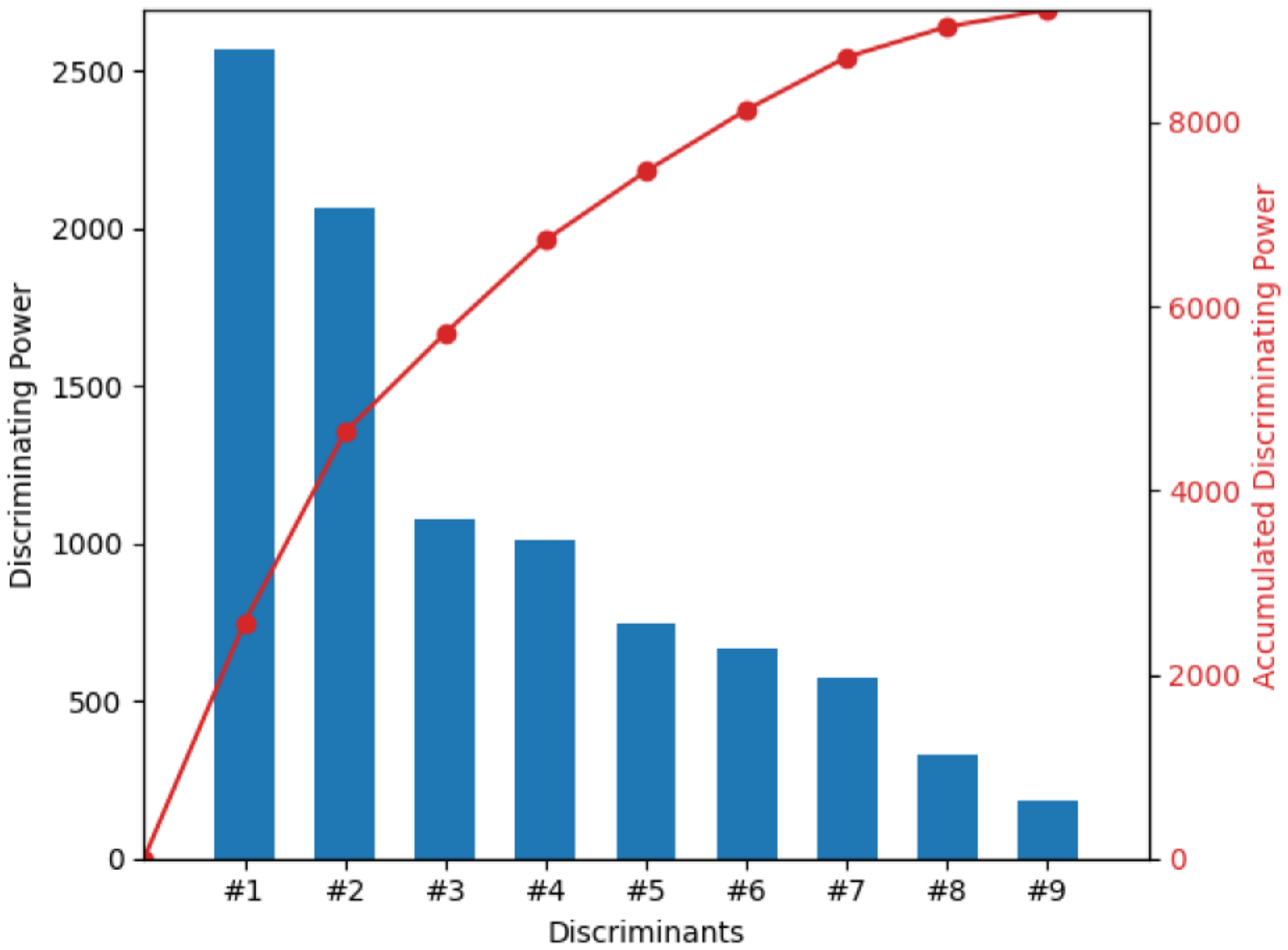}
    \caption{with pushing objective}
    \label{fig:discrimwithpush}
\end{subfigure}
~
\begin{subfigure}{0.475\linewidth}
    \includegraphics[clip, trim=0 0 0 0,width=\linewidth]{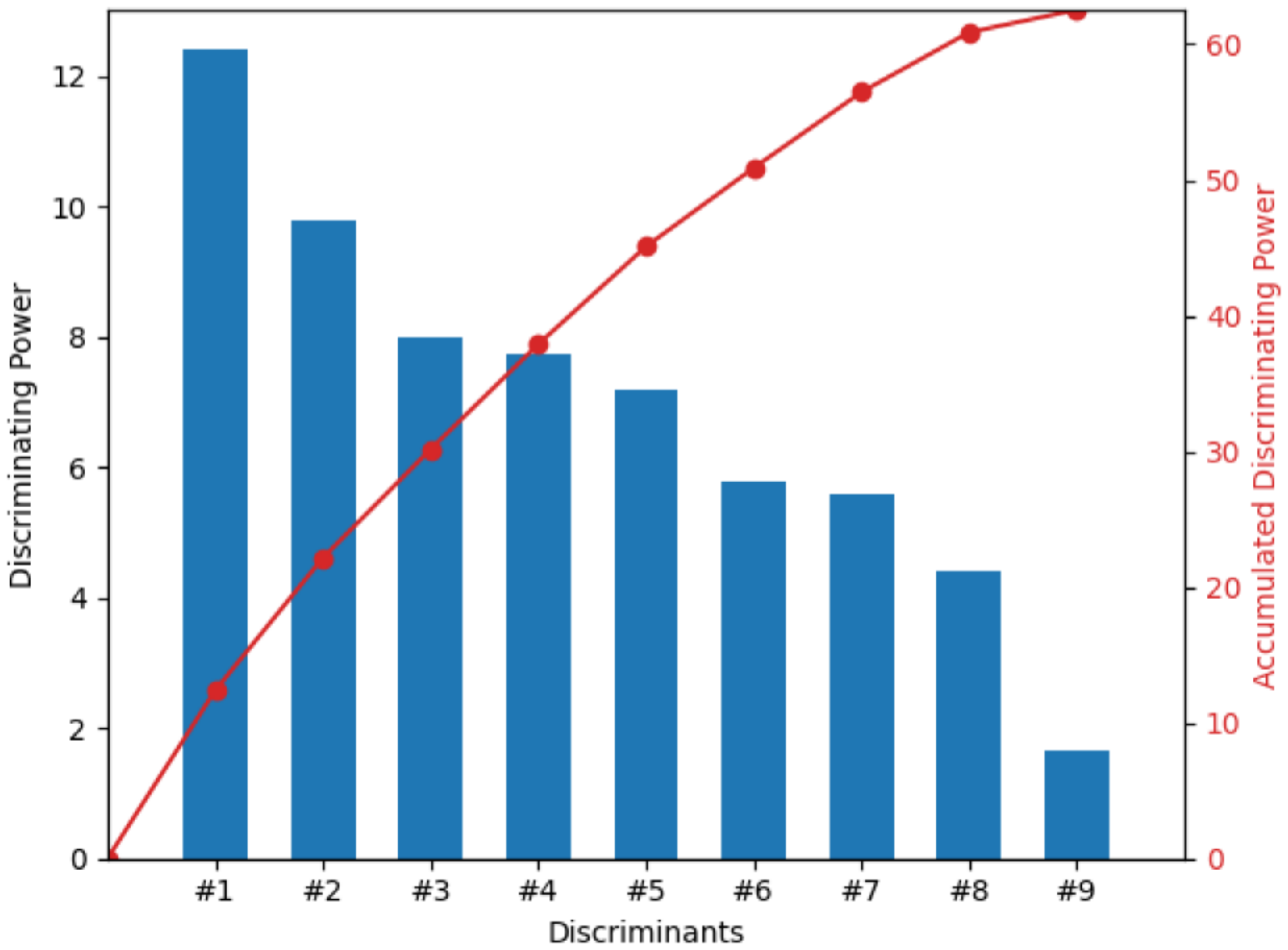}
    \caption{without pushing objective}
    \label{fig:discrimwithoutpush}
\end{subfigure}
\caption{Top nine discriminants after training (a) with and (b) without our pushing objective. The horizontal axis represents the nine top discriminants and the left vertical axis indicates their corresponding discriminating power ($v_j$ in Eq.~\ref{eq:geigenvaluepplusp} and Eq.~\ref{eq:vjp}). The right vertical axis and the curve in red denote the accumulated discriminating power.}
\label{fig:mnistdiscriminants}
\vspace{-0.05in}
\end{figure}
Figure~\ref{fig:mnistdiscriminants} shows the top nine discriminants after training with and without our pushing objective. As expected, the discriminating power ($v_j$ in Eq.~\ref{eq:geigenvaluepplusp}), is improved with our deep LDA pushing by two orders of magnitude. Also, the distribution after pushing is more spiky and, in terms of proportion, more discriminating power is pushed to the large discriminants on the left. This can also be seen from the red accumulative discriminating power curve in Fig.~\ref{fig:discrimwithpush} and~\ref{fig:discrimwithoutpush}. The first two discriminants count for about one third of the total discriminating power in the no-push case (Fig.~\ref{fig:discrimwithoutpush}) while this number increases to 50\% for the case with our pushing objective (Fig.~\ref{fig:discrimwithpush}). It means that when pruning, we can throw away more neuron dimensions while still maintaining enough discriminating power.
In this simple example, all neurons other than the top nine are put to dormant (with 0 discriminating power) after pushing while there are more active neurons
in the no-push case.
In our experiments, after reducing the original network size from 4.0M parameters to only 38.6K parameters, we can still maintain comparable test accuracy to the original.

\subsection{CIFAR10}

Please refer to~\cite{krizhevsky2009learning} for dataset details. We set aside the first 10,000 images in the training set for validation.

\subsubsection{Accuracy change v.s. parameters pruned}

\begin{figure}
\centering 
\begin{subfigure}{0.75\linewidth}
    \includegraphics[width=\linewidth]{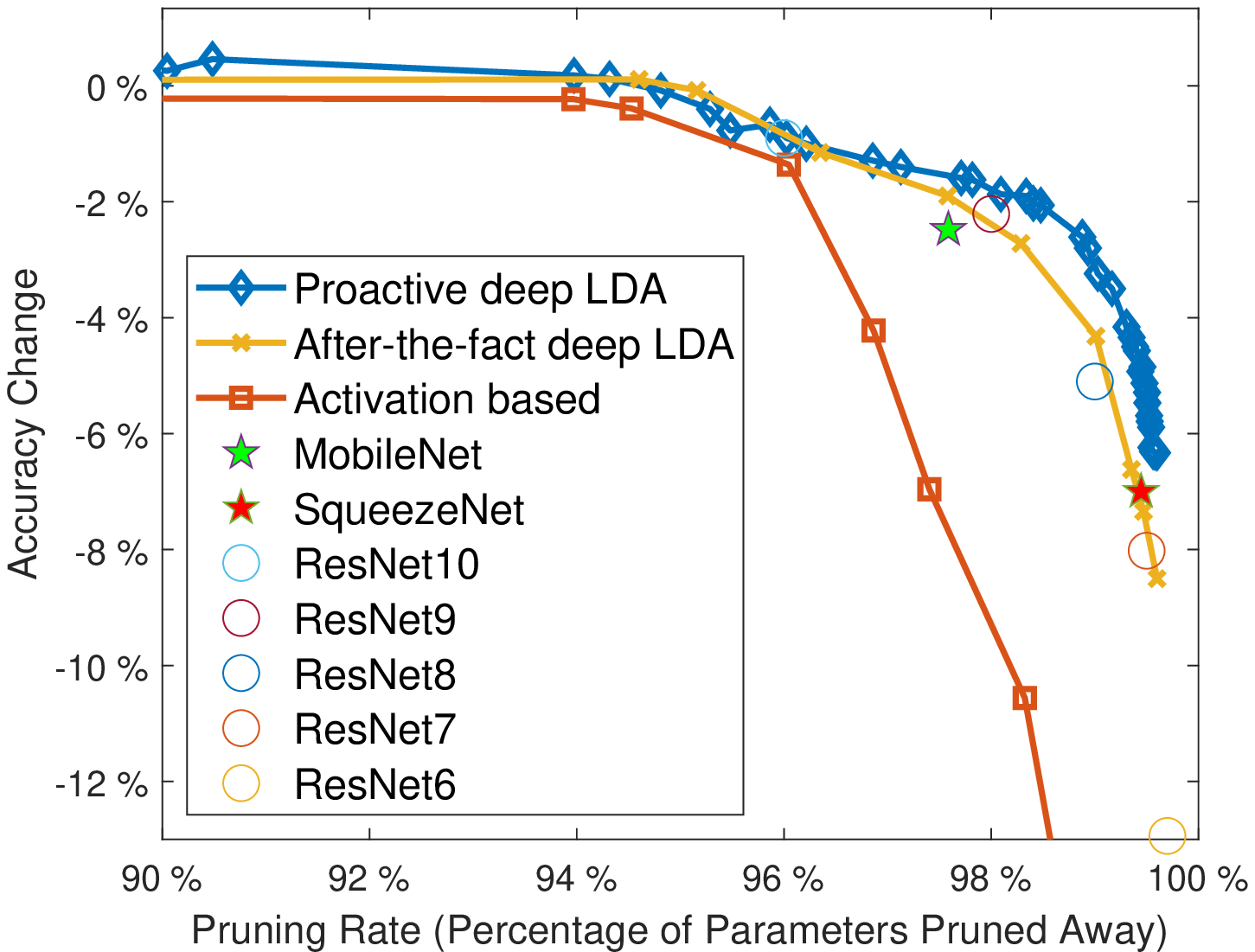}
    \caption*{CIFAR10, base: 95.19\%, after pushing: 95.72\%}
\end{subfigure}
\caption{Accuracy change v.s. parameter pruning rate on CIFAR10. In addition to our proactive deep LDA pruning, we add after-the-fact deep LDA pruning~\citep{tian2021}, activation-based pruning (as in~\cite{molchanov2016pruning}), MobileNet~\citep{howard2017}, SqueezeNet~\citep{iandola2016}, and tiny ResNets (details in Table~\ref{tab:tinyresnets}) for comparison. Small pruning rates are skipped where accuracy changes little. The original base and competing fixed models are pre-trained on ImageNet.}
\label{fig:cifar10paramcomparison}
\end{figure}

The number of parameters is an important measure of complexity. Model size directly determines where the model/features reside, whether they can fit to different levels of caches and memories. It is memory accesses (rather than operations) that dominantly influence energy and latency \citep{horowitz20141}.
In this experiment, we start with a VGG-16 model pre-trained on ImageNet. Cross-entropy loss with $L_2$ regularization leads to a validation accuracy of 95.19\% on CIFAR10. In addition to aligning discriminants with neuron dimensions, our deep LDA pushing objective helps improve the accuracy to 95.72\% without pruning. Figure~\ref{fig:cifar10paramcomparison} illustrates the change of accuracy with respect to parameter pruning rate (the percentage of parameters discarded from the original model). We focus on high pruning rates where the accuracy changes fast with the decrease of parameters. That said, it is worth noting that among the few small pruning rates investigated, a pruned model with 118M parameters enjoys an even better accuracy (96.01\%) than both the original model and the pushed one.
For comparison, we add after-the-fact deep LDA pruning~\cite{tian2021} and activation-based filter pruning (as mentioned in~\cite{molchanov2016pruning}), which treats filter importance as average activation magnitudes/variances within a filter. Also, we compare our method with some popular compact fixed nets, i.e., MobileNet~\cite{howard2017}, SqueezeNet~\cite{iandola2016}, and tiny ResNets. Here, tiny ResNets refer to residual nets with shallow depths. In this experiment, we test ResNet6 - ResNet10. Their detailed configurations are shown in Table~\ref{tab:tinyresnets}.

\begin{table}
  \begin{center}
    \begin{tabular}{cc}
      \hline
      Name & Configuration \\
      \hline
      ResNet6 & i64-c128 \\
      ResNet7 & i64-c128-1c256 \\
      ResNet8 & i64-c128-c256 \\
      ResNet9 & i64-c128-c256-1c512 \\
      ResNet10 & c64-c128-c256-c512 \\
      \hline
    \end{tabular}
  \end{center}
  \caption{Tiny ResNets used as comparison in our experiments on CIFAR10. The dash sign `-' separates different stages. As defined in~\cite{he2015}, there are two types of residual modules, i.e., identity module and convolutional module where $1\times1$ filters are employed on the shortcut path to match dimension. Only depth-2 modules are used here. In this table, `i' stands for depth-2 identity block and `c' represents depth-2 convolutional block. The number follows `i' or `c' indicates the number of filters within each conv layer in that module. In addition to residual modules, we adopt the same stem layers as in~\cite{he2015}.}
  \label{tab:tinyresnets}
\end{table}

As we can see from the results, our proactive-deep-LDA pruning, generally speaking, enjoys higher accuracy than the other two pruning approaches and the compact nets at similar complexities. The gaps are more obvious at high pruning rates, especially between activation-based pruning and our proactive deep LDA pruning. This performance difference implies that strong activation does not necessarily indicate high final classification utility. It is possible that some strong yet irrelevant activation skews or misleads the data analysis at the top of the network. Compared to after-the-fact deep LDA pruning, the proactive deep LDA pruning proposed in this paper enjoys better performance. The reason is that although after-the-fact deep LDA is capable of capturing final class separation utility, useful and useless components may already be mixed in the given pre-trained model, and it is hard to trim one without influencing the other.
The performance differences are small at low pruning rates, perhaps because even when `useful' feature components are discarded, the network can recover such or similar features through re-training when pruning rates are low. This `learning to repair' ability via re-training gradually declines when the network capacity becomes small.
Furthermore, even though ResNet is currently one of the most successful deep nets, stacking a few residual modules with random numbers of filters only leads to suboptimal performance compared to the proposed proactive deep LDA pruning. In Figure~\ref{fig:cifar10paramcomparison}, our deep LDA-pushed-and-pruned models beat tiny ResNets at most similar complexities. This indicates the necessity of informed pruning over architecture hand-engineering with human expertise.

\subsubsection{Layerwise complexity}

\begin{figure*}[!htp]
\centering
\begin{subfigure}{0.31\linewidth}
    \includegraphics[width=\linewidth]{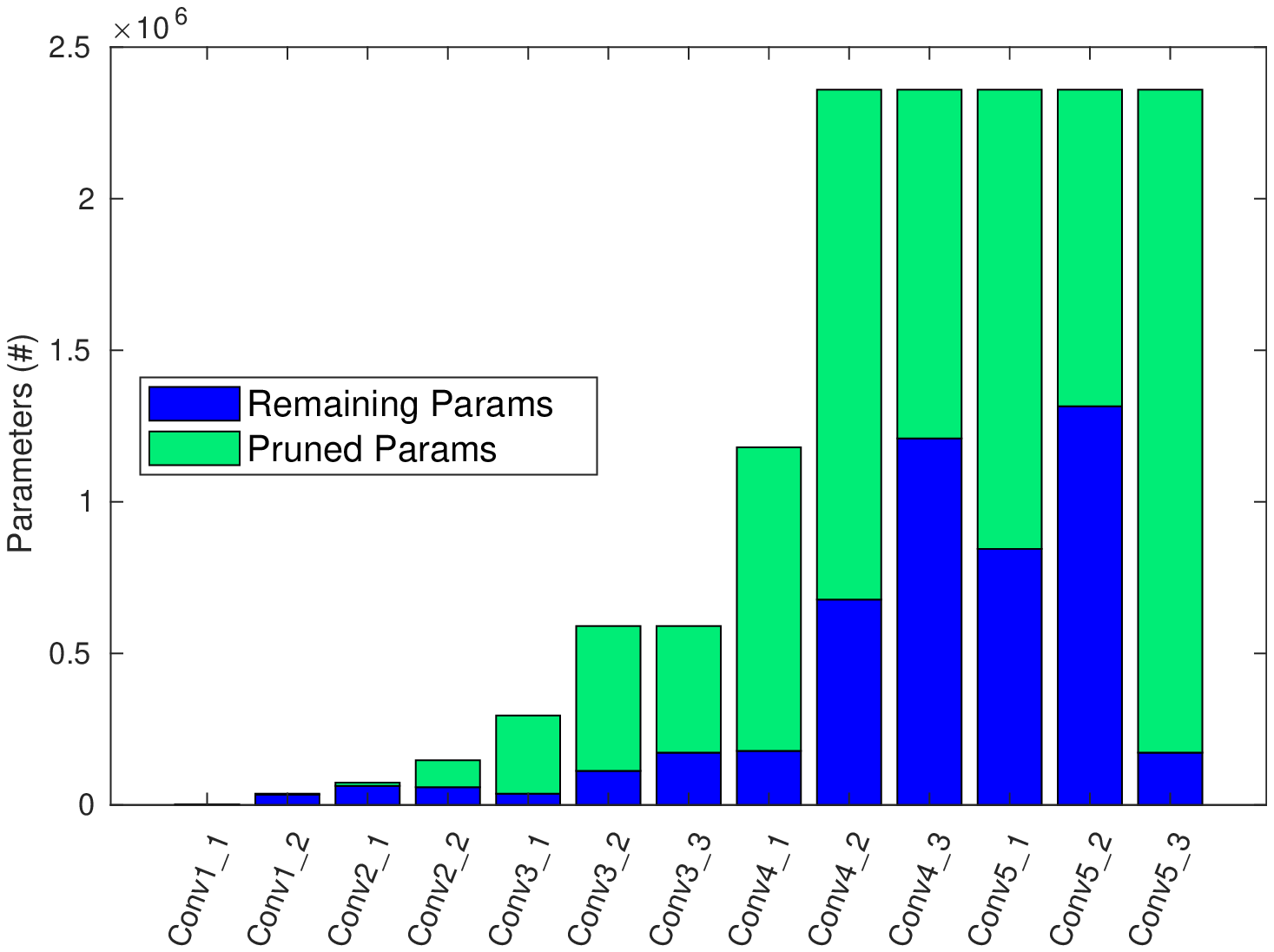}
    \caption{Conv param}
     \label{fig:paramcifar10layerwise}
\end{subfigure}
~ 
\begin{subfigure}{0.31\linewidth}
    \includegraphics[width=\linewidth]{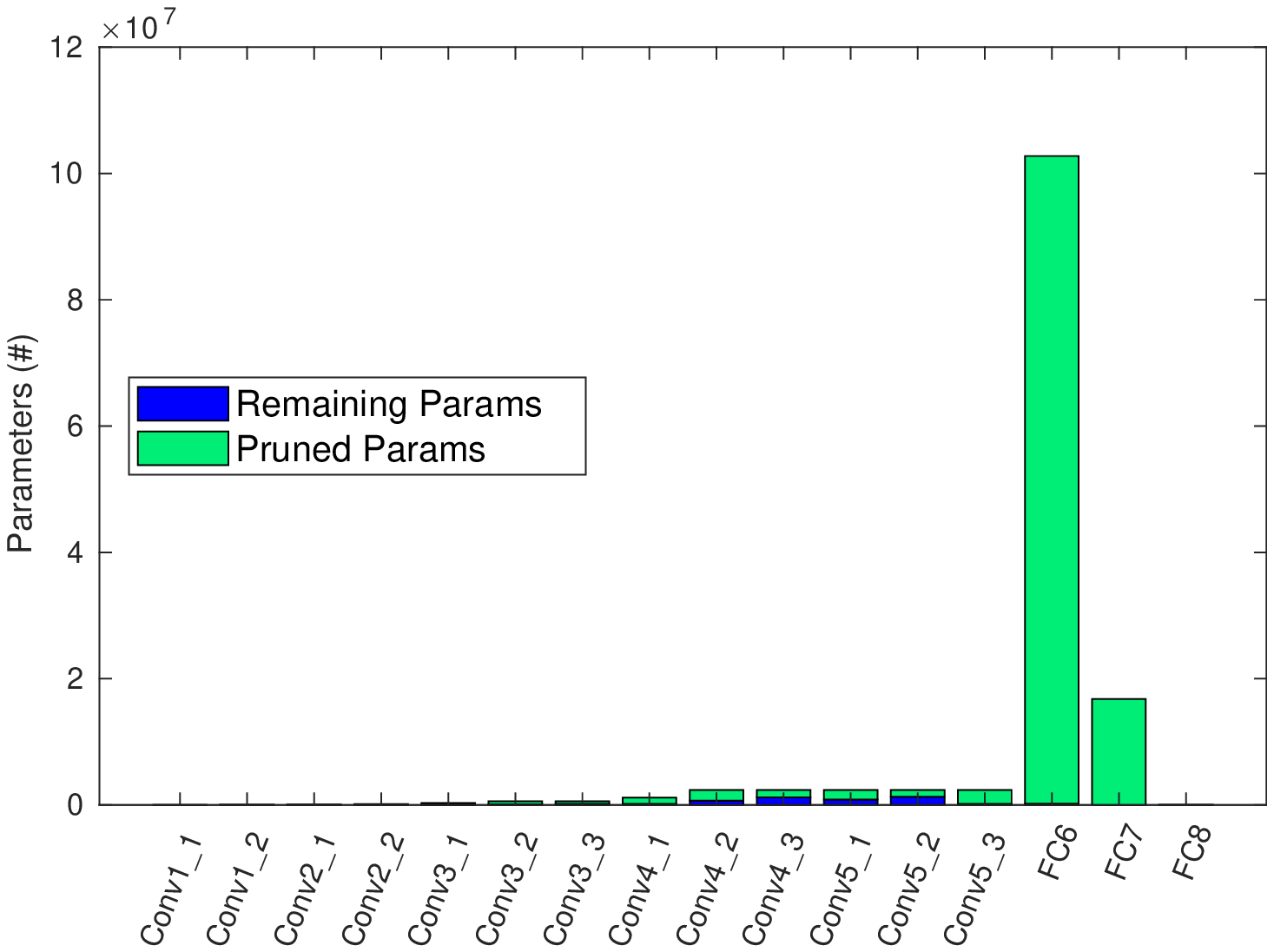}
    \caption{Param savings}
    \label{fig:paramcifar10layerwise_FC}
\end{subfigure}
~
\begin{subfigure}{0.31\linewidth}
    \includegraphics[width=\linewidth]{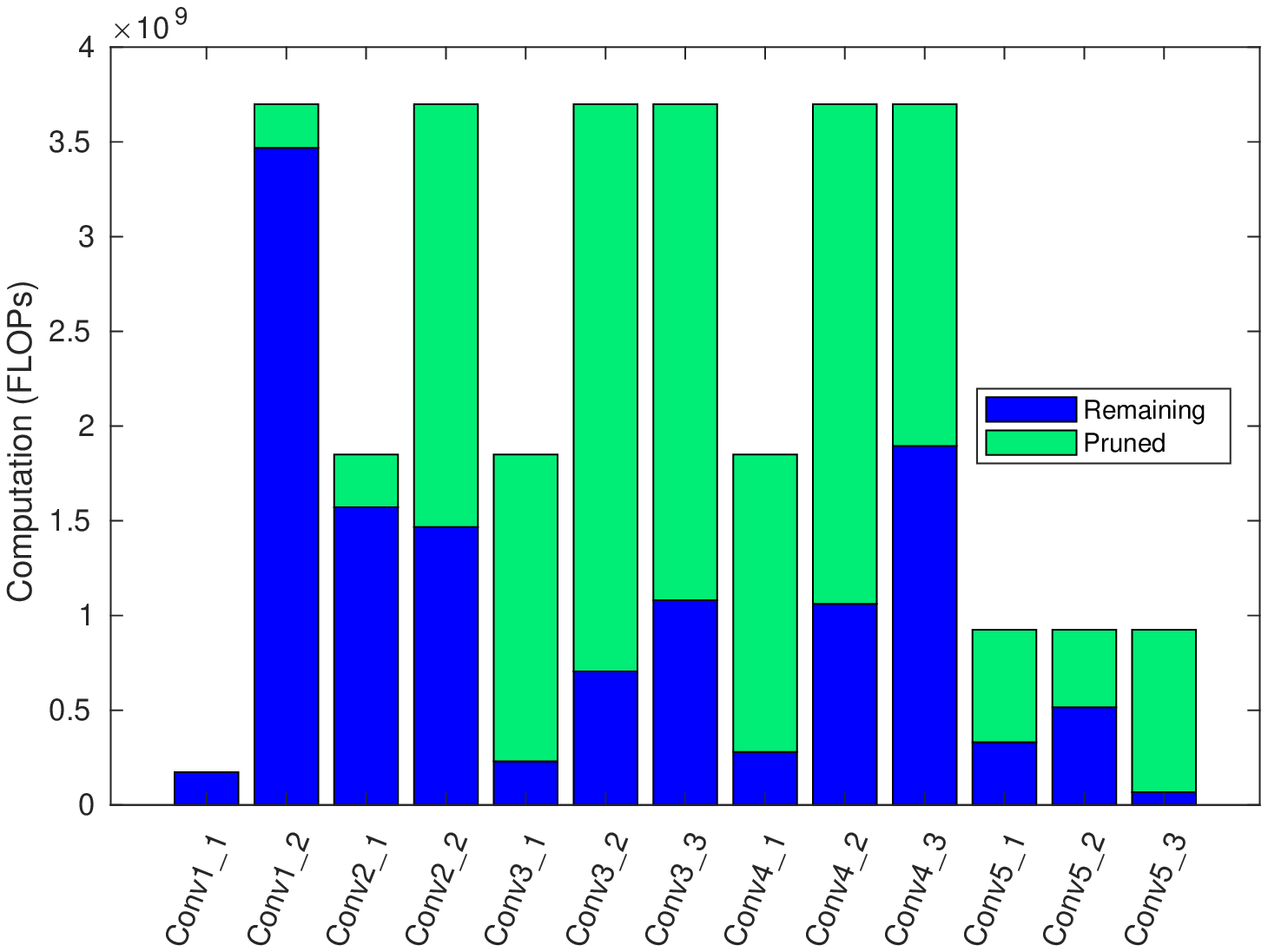}
    \caption{FLOP savings}
     \label{fig:flopcifar10layerwise}
\end{subfigure}
\caption{Layerwise complexity reductions (CIFAR10, VGG16). Green: pruned, blue: remaining. We add a separate parameter analysis for conv layers because FC layers dominate the model size. Since almost all computations are in the conv layers, only conv layer FLOPs are demonstrated.}
\label{fig:layerwisecifar10}
\end{figure*}

Figure~\ref{fig:layerwisecifar10} demonstrates the layerwise complexity of our smallest pruned model that maintains comparable accuracy to the original VGG-16. FC layers dominate the original net size, while almost all computation comes from conv layers. According to the results, most parameters and computations have been thrown away in the layers except for the first three layers that capture commonly useful patterns.

\subsection{ImageNet}\label{sec:proactiveldaimagenet}

In this subsection, we demonstrate our `grow-push-prune' pipeline's efficacy on the ImageNet (ILSVRC12) dataset~\cite{ILSVRC15}. The dataset is widely used for benchmarking algorithms in computer vision and machine learning.
In our experiment, all the images are pre-resized to 256$\times$256. During training, the images are randomly cropped to 224$\times$224 and randomly mirrored about the vertical axis. Like many previous works, we report accuracy change on the validation set (no test set labels are publicly available) and only the center crop is used.

\subsubsection{Base net growing and pushing}
Table~\ref{tab:deepinceptionnets} shows some models encountered in the growing step using the basic Inception module on ImageNet. The accuracy in Table~\ref{tab:deepinceptionnets} is Top-1 accuracy using only one center crop. The name Inception-N means the net is $N$-layer deep (only conv and fully-connected layers are considered).

\begin{table*}[htp]
  \begin{center}
    \begin{tabular}{cccccc}
      \hline
      Name & Modules & Stage size & Parameters & FLOPs & Accuracy\\
      \hline
      InceptionV1 & 9 & (2,5,2) & 6.7M & 3.0B & 70.64\% \\ 
      Inception-34 & 10 & (3,5,2) & 7.1M & 3.7B & 71.12\% \\ 
      Inception-37 & 11 & (3,5,3) & 8.6M & 3.8B & 71.75\% \\ 
      Inception-40 & 12 & (4,5,3) & 9.0M & 4.5B & 71.97\% \\ 
      Inception-43 & 13 & (4,6,3) & 10.0M & 4.9B & 72.03\% \\ 
      Inception-46 & 14 & (4,7,3) & 11.0M & 5.3B & 73.45\% \\ 
      Inception-49 & 15 & (4,7,4) & 12.5M & 5.4B & 73.51\% \\ 
      Inception-52 & 16 & (4,7,5) & 14.0M & 5.6B & 73.69\% \\ 
      Inception-55 & 17 & (4,8,5) & 15.0M & 6.0B & 73.91\% \\ 
      Inception-58 & 18 & (5,8,5) & 15.5M & 6.6B & 74.27\% \\ 
      Inception-61 & 19 & (6,8,5) & 15.9M  & 7.3B & 74.20\% \\ 
      Inception-64 & 20 & (7,8,5) & 16.3M & 8.0B & 74.42\% \\ 
      Inception-67 & 21 & (7,8,6) & 17.8M & 8.1B & 74.58\% \\ 
      Inception-70 & 22 & (7,8,7) & 19.3M & 8.3B & 74.54\% \\ 
      Inception-73 & 23 & (8,8,7) & 19.8M  & 8.9B & 74.64\% \\ 
      Inception-76 & 24 & (8,8,8) & 21.3M & 9.1B & 74.60\% \\ 
      Inception-79 & 25 & (8,8,9) & 22.8M & 9.2B & 74.59\% \\ 
      Inception-82 & 26 & (9,8,9) & 23.2M  & 9.9B & 74.77\% \\ 
      Inception-85 & 27 & (9,8,10) & 24.7M & 10.1B & 74.60\% \\ 
      \textbf{Inception-88} & \textbf{28} & \textbf{(10,8,10)} & \textbf{25.1M} & \textbf{10.7B} & \textbf{75.01\%} \\
      Inception-91 & 29 & (10,8,11) & 26.6M & 10.9B & 74.71\% \\
      \hline
    \end{tabular}
  \end{center}
  \caption{Deep Inception net examples encountered in the base net growing process on ImageNet. The accuracy here indicates Top-1 accuracy using only one center crop. The name Inception-N means the net is $N$-layer deep (only conv and fully-connected layers are considered). The stage size column shows module numbers across the three stages. M=$10^6$, B=$10^9$.}
  \label{tab:deepinceptionnets}
\end{table*}

As can be seen from the results, we can obtain more accuracy by simply stacking more modules to the basic InceptionNet stages (Algorithm~\ref{alg:greedygrow}).
Specifically, we would like to introduce Inception-88 that achieves comparable accuracy (75.01\%) to ResNet-50 (74.96\%\footnote{Unlike the ResNet-50 achieving 76\% in Tensorflow, no bounding box info is used in any of our models. Only 1-center crop is used for validation.}) at a slightly smaller complexity. Even better, there are no hard-coded dimension alignment by human experts, which would limit the room for pruning.
The depths of the three stages in Inception-88 are respectively 30, 24, and 30. Beyond Inception-88, accuracy first drops slightly before increasing slowly with increasing module number. This is also similar to the ResNet-50 case where 19M more parameters (ResNet-101) only increase the accuracy by less than 1\%~\citep{he2016}.
Apart from increasing capacity and accuracy, this growing step offers more wiggle/adjustment room for the net to re-organize utility during the pushing step. 
Our pushing step increases Inception-88's accuracy to 75.2\%, in addition to compressing and aligning utility with latent neuron dimensions.

\subsubsection{Accuracy change vs. pruning rate}

In the final pruning step, we strip off the separated unnecessary complexities. In this grow-push-prune pipeline, bottom-up search and top-down search are combined. 

\begin{figure*}
\centering
\begin{subfigure}{\linewidth}
    \includegraphics[clip, trim=1.6in 0.2in 0.45in 0,width=\linewidth]{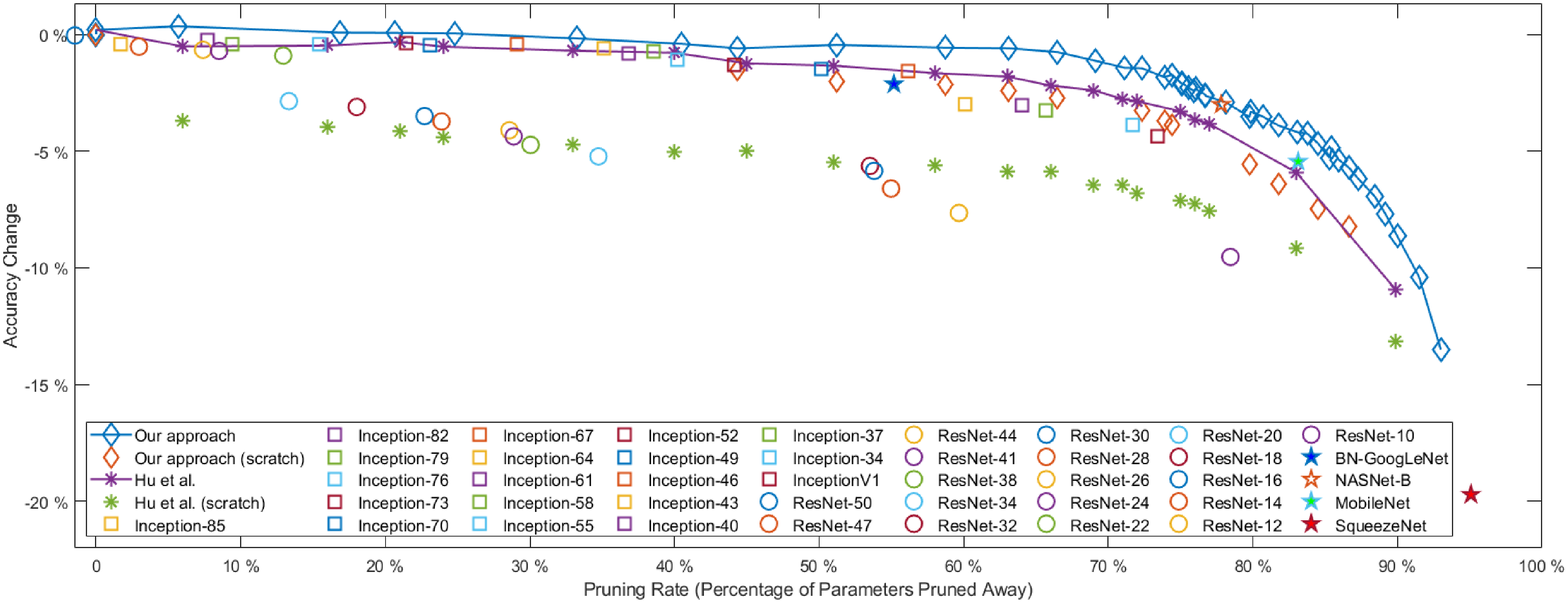}
    \caption*{ImageNet, base accuracy: 75.01\%, after pushing: 75.20\%}
\end{subfigure}
\caption{Accuracy change vs. parameter pruning rate on ImageNet. In addition to our deep LDA push-and-prune method (blue), we add our grown deep Inception nets, various ResNets
, network slimming \citep{hu2016network}, and popular fixed compact nets for comparison. 
There are two accuracies at 0 pruning rate. The upper one represents Inception-88 trained with our deep LDA push objective added. The negative pruning rate of ResNet-50 indicates its larger size than Inception-88. Our derived nets trained from scratch (red diamonds) mark the beginning of each iteration for our approach. NASNet-B architecture is from~\cite{zoph2017learning} and trained from scratch in the same experimental setting as ours. 
More iterations (smaller pruning steps) may further improve the pruned models' accuracy.
}
\label{fig:proactiveldaimagenet}
\end{figure*}

Figure~\ref{fig:proactiveldaimagenet} demonstrates the influence of our pruning on accuracy. As can be seen, comparable accuracy can be maintained even after about 70\% parameters are discarded. At the pruning rate of approximately 6\%, a pruned model achieves an accuracy of 75.36\% which is higher than that of both unpruned versions.
\begin{table*}
  \begin{center}
    \begin{tabular}{ll}
      \hline
      Name & Configuration \\
      \hline
      ResNet6 & i64-c128 \\
      ResNet7 & i64-c128-1c256 \\
      ResNet8 & i64-c128-c256 \\
      ResNet9 & i64-c128-c256-1c512 \\
      ResNet10 & c64-c128-c256-c512 \\
      ResNet12 & (c64,~i64)-c128-c256-c512 \\
      ResNet18 & (c64,~i64)-(c128,~i128)-(c256,~i256)-(c512,~i512) \\
      ResNet20 & (c64,~i64)-(c128,~i128)-(c256,~i256)-(c512,~i512,~i512) \\
      ResNet22 & (c64,~i64)-(c128,~i128)-(c256,~i256,~i256)-(c512,~i512,~i512) \\
      ResNet24 & (c64,~i64)-(c128,~i128,~i128)-(c256,~i256,~i256)-(c512,~i512,~i512) \\
      ResNet26 & (c64,~i64,~i64)-(c128,~i128,~i128)-(c256,~i256,~i256)-(c512,~i512,~i512) \\
      ResNet28 & (c64,~i64,~i64)-(c128,~i128,~i128)-(c256,~i256,~i256,~i256)-(c512,~i512,~i512) \\
      ResNet30 & (c64,~i64,~i64)-(c128,~i128,~i128,~i128)-(c256,~i256,~i256,~i256)-(c512,~i512,~i512) \\
      ResNet32 & (c64,~i64,~i64)-(c128,~i128,~i128,~i128)-(c256,~i256,~i256,~i256,~i256)-(c512,~i512,~i512) \\
      ResNet34 & (c64,~i64,~i64)-(c128,~i128,~i128,~i128)-(c256,~i256,~i256,~i256,~i256,~i256)-(c512,~i512,~i512) \\
      \hline
      ResNet38 & (C64,~I64,~I64)-(C128,~I128,~I128)-(C256,~I256,~I256)-(C512,~I512,~I512) \\
      ResNet41 & (C64,~I64,~I64)-(C128,~I128,~I128)-(C256,~I256,~I256,~I256)-(C512,~I512,~I512) \\
      ResNet44 & (C64,~I64,~I64)-(C128,~I128,~I128,~I128)-(C256,~I256,~I256,~I256)-(C512,~I512,~I512) \\
      ResNet47 & (C64,~I64,~I64)-(C128,~I128,~I128,~I128)-(C256,~I256,~I256,~I256,~I256)-(C512,~I512,~I512) \\
      ResNet50 & (C64,~I64,~I64)-(C128,~I128,~I128,~I128)-(C256,~I256,~I256,~I256,~I256,~I256)-(C512,~I512,~I512) \\
      \hline
    \end{tabular}
  \end{center}
  \caption{ResNets used as comparison in our experiments on ImageNet. The dash sign `-' separates different stages. As defined in~\cite{he2015}, there are two types of residual modules, i.e., identity module and convolutional module where $1\times1$ filters are employed on the shortcut path to match dimension. Here, `i' stands for depth-2 identity block, `c' represents depth-2 convolutional block, `I' stands for depth-3 identity block, and `C' represents depth-3 convolutional block. The number follows `i', `c', `I', or `C' indicates the number of filters within each conv layer in that module. Parentheses are used to group multiple modules in a stage. In addition to residual modules, we adopt the same stem layers as in~\cite{he2015}.}
  \label{tab:resnets}
\end{table*}
For comparison, we add to Figure~\ref{fig:proactiveldaimagenet} the results of the deep Inception nets derived from the growing step, a range of ResNets, models from network trimming~\citep{hu2016network}, and popular compact nets, including SqueezeNet~\citep{iandola2016}, MobileNet~\citep{howard2017}, NASNet-B~\citep{zoph2017learning}, BN-GoogLeNet\footnote{It is not just GoogLeNet + batchnorm. There are more architectural changes to GoogLeNet which we do not include in our grown deep Inception nets.}~\citep{ioffe2015batch}. We also include the results of training some pruned architectures from scratch (no weights are inherited). 
The detailed configurations of the ResNets used for comparison are shown in Table~\ref{tab:resnets}. Starting from ResNet-50, a module is iteratively removed from the longest stage to produce smaller ResNets.
When two stages have the same number of modules, we follow a bottom-to-top order to choose which module to remove (until ResNet18). From ResNet-50 to ResNet-38, the residual modules are of depth 3. From ResNet-34 downwards, each module has a maximum depth of 2. The depth-2 and depth-3 residual modules are defined the same as in~\cite{he2015}.

According to Figure~\ref{fig:proactiveldaimagenet}, we can see that our compact models pushed-and-pruned from Inception-88 outperform smaller deep Inception nets grown, the resnets, nets trimmed using~\cite{hu2016network}, and the fixed compact nets.
The pruned models, by \cite{hu2016network} and our approach, achieve better performance compared to training the same architectures from scratch. This highlights the value of the knowledge acquired by and transferred from the larger grown-pushed base model in the form of weights. That said, even when trained from scratch, our pruned nets still attain satisfactory accuracy and beat many others, including those produced by network trimming~\citep{hu2016network} and then trained from scratch. It means that, besides the weights, there is some value in our derived architectures themselves.

Also, Figure~\ref{fig:proactiveldaimagenet} reveals that our grown series of deep Inception nets outperform the residual structures at similar complexities. 
As far as we know, this is the first time that a range of basic Inception structures are fairly compared against residual structures on the same input, at least in the complexity range we investigated. 
Another advantage of these deep Inception nets over the residual structures is that the former does not need to enforce the output dimensions of a module's branches to be the same. To our best knowledge, there is no theoretical justification why different branches need to have exactly the same dimension. It is possible that there is more information lying on the 3$\times$3 scale than others. As a result, in a ResNet module, the output dimensions will most likely not agree after a filter-level pruning based on importance (unless the agreement is enforced at the expense of more complexity). In this sense, deep Inception nets are more conducive to pruning approaches in general.

Compared to the three fixed nets shown as five-pointed stars in Fig.~\ref{fig:proactiveldaimagenet}, the proposed pipeline not only achieves better accuracy at similar complexities but also offers a wide range of compact models for different accuracy and complexity requirements.

In contrast to expensive NAS approaches that train numerous architecture samples separately or based on ad hoc relations, our top-down search only needs to sample architectures along the direction aligned with task utility (e.g., $10^4$ samples in NASNets \cite{zoph2017learning} vs. $10^1$ in ours). Unlike methods that `predict' post-retraining performance, we can afford to fully retrain sampled architectures. Additionally, useful parameters inherited from the previous base make the sample architecture retraining process converge very fast. Usually, it only takes a few epochs to achieve accuracy within 5\% from that of the fully trained.

With extra training data, more computing budget (more pruning iterations), and some tricks previously mentioned, the accuracy reported may be further improved. That said, achieving the best accuracy possible with non-architectural tricks is not the focus of this paper and is deferred to future work.

\subsubsection{Layerwise complexity}

\begin{figure*}
\centering
\includegraphics[clip, trim=2.0in 0.9in 1.65in 0.55in,width=0.9\linewidth]{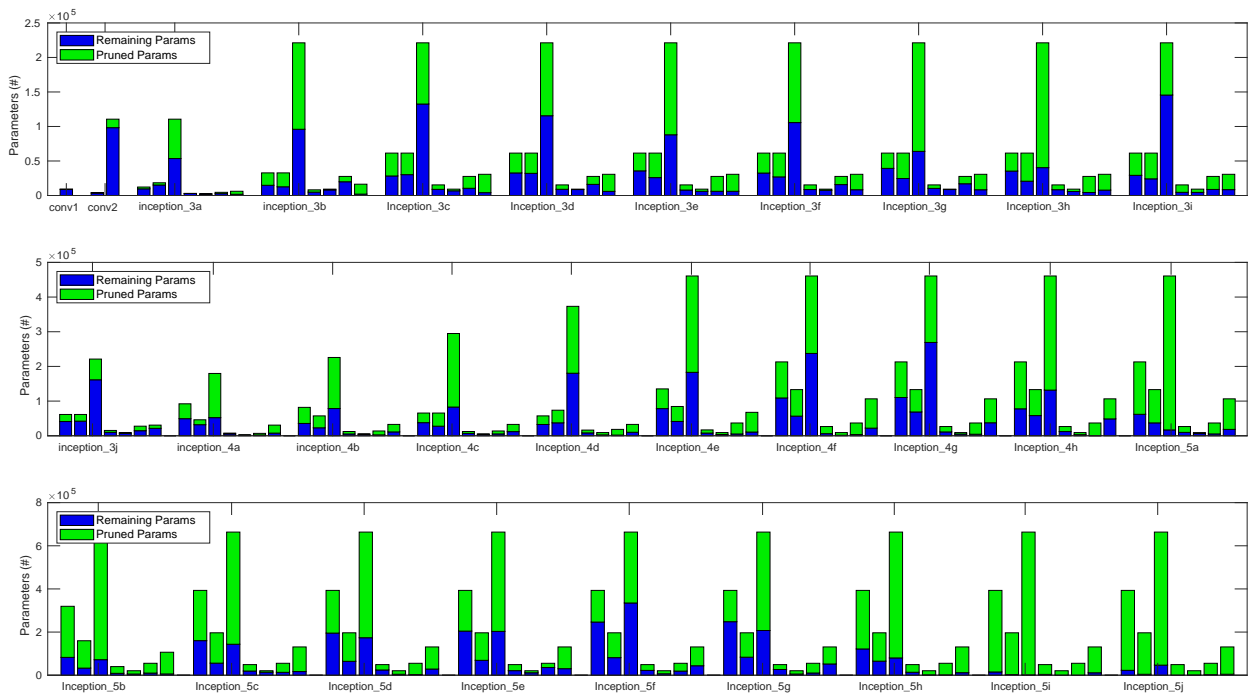}
\caption{Layerwise parameter reductions of the grown Inception-88 on ImageNet. From left to right, the conv layers in a Inception module are ($1\times1$), ($1\times1$, $3\times3$), ($1\times1$, $3\times3$ a, $3\times3$ b), ($1\times1$ after pooling). Green: pruned, blue: remaining. Due to the large network depth, the layer-wise parameter complexity figure is displayed in three rows. conv2 includes a dimension reducing layer in front (notation skipped because of space limit).}
\label{fig:layerwiseparamimagenetp}
\end{figure*}

\begin{figure*}
\centering
\includegraphics[clip, trim=2.0in 0.9in 1.65in 0.55in,width=0.9\linewidth]{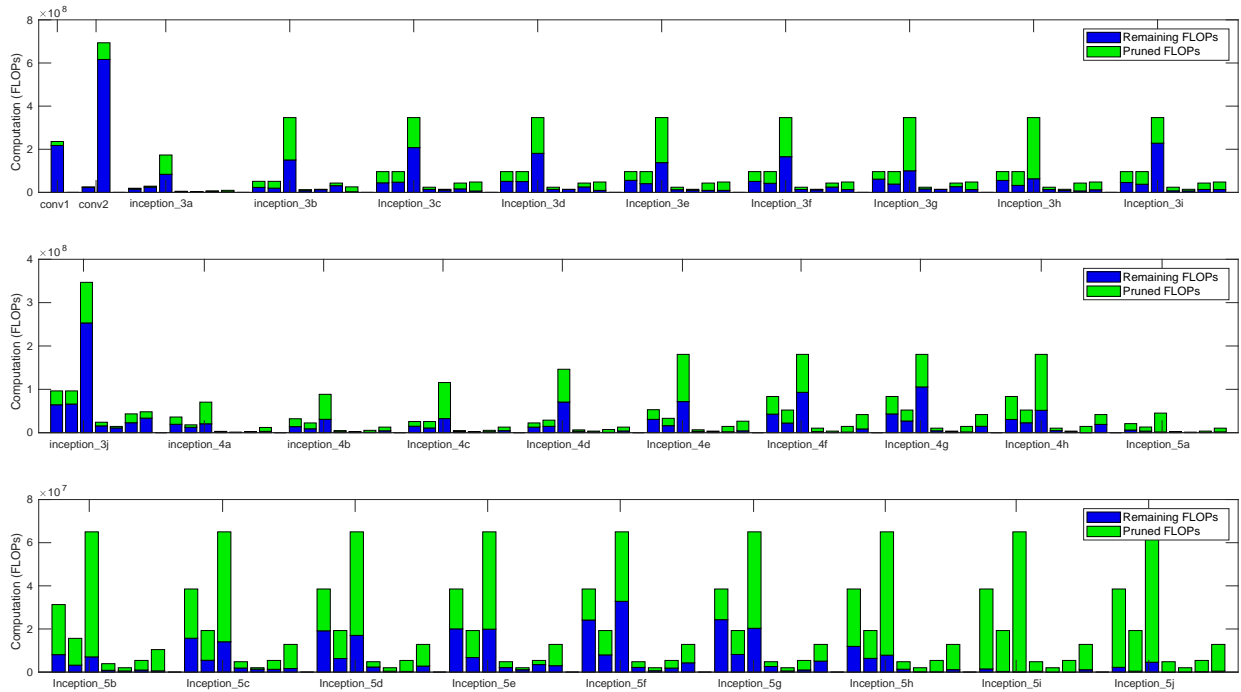}
\caption{Layerwise FLOPs reductions of the grown Inception-88 on ImageNet. From left to right, the conv layers in a Inception module are ($1\times1$), ($1\times1$, $3\times3$), ($1\times1$, $3\times3$ a, $3\times3$ b), ($1\times1$ after pooling). Green: pruned, blue: remaining. Due to the large network depth, the layer-wise FLOPs complexity figure is displayed in three rows. conv2 includes a dimension reducing layer in front (notation skipped because of space limit).}
\label{fig:layerwiseflopsimagenetp}
\end{figure*}

Figure~\ref{fig:layerwiseparamimagenetp} and~\ref{fig:layerwiseflopsimagenetp} visualize layer-wise parameter and FLOPs complexity reduction results of our pruning on the `grown-pushed' Inception-88 model (with comparable accuracy maintained). From left to right, the conv layers within a Inception module are (1$\times$1), (1$\times$1,3$\times$3), (1$\times$1,3$\times$3a,3$\times$3b), (1$\times$1 after pooling) layers.

According to Figure~\ref{fig:layerwiseparamimagenetp} and~\ref{fig:layerwiseflopsimagenetp}, most parameters and computations over the layers are pruned away, and different types of filters are pruned differently depending on the abstraction level and the scales where more task utility lies.
As anticipated, the pruning rates of the first few layers, which capture commonly useful primitive patterns, are low. Almost all of the parameters and FLOPs are pruned away in the last two modules, which can be regarded as an indicator that the depth is large enough (at least locally). This is in agreement with our observation at the growing step that adding one or two more modules to the Inception-88 net does not help much.

Interestingly, while the deep Inception net was greedily grown to achieve the highest accuracy locally, there are still massive redundant and useless structures over the layers. That is to say, at the growing step, each time we stacked one more module in the attempt to gain more accuracy, we simultaneously added more useless structures due to the ad hoc filter numbers used. Those useless structures cannot be effectively aligned with task utility even after training and can thus be discarded. The large pruning rates over the layers highlight our approach's advantage over architecture hand-engineering with ad-hoc filter numbers.

\section{Discussion and Future directions}

Deep discriminant analysis, a non-linear generalization of LDA, is able to pick up high-order moments/statistics embedded in the complex raw data space with the help of deeply learned transformation. As a future direction, we plan to extend the idea of deep discriminant analysis and reduction to visual detection tasks. Visual detection involves both class separation and localization, which we believe are closely related. An accurate localization is based on correctly identifying distinguishing features for each class. We plan to design a location-aware variant of deep discriminant analysis to quantify the detection utility.

Also, it would be of great interest to investigate our deep-discriminant-based pruning's influence on model robustness. We believe that two important causes of adversarial vulnerability are over-fitting and the model's inadequacy to accurately capture the task demands. Our hypothesis is that adversarial attacks can trigger interfering features that are not aligned with task demands. The more such task-irrelevant features a model has, the higher the chance it will be hit by adversarial attacks and noises. By discarding irrelevant structures, we expect to mitigate overfitting and remove interfering parameters, thus decreasing the chance of the model falling victim to irrelevant factors in the image space.

Furthermore, the proposed growing strategy based on Inception modules can potentially offer more plasticity for transfer learning and domain adaptation tasks, which will be investigated in our future work.

\section{Conclusion}

In this paper, instead of relying on a pre-trained model, we have proposed a proactive pruning approach for compact architecture search based on deep discriminant analysis. The approach follows a two-step procedure in iterations: (1) through learning, it proactively maximizes and unravels twisted threads of deep discriminants, condenses and pushes them into alignment with a subset of neurons; (2) after useful features are separated from others, the second pruning step simply throws away the useless or even harmful components over the layers based on deconv tracing.
In addition, we explore to grow the base model for tasks requiring larger capacity.
We demonstrate our methods' efficacy on the MNIST, CIFAR10, and ImageNet datasets. 
By growing from the basic InceptionV1 to an 88-layer-deep Inception variant, we show that deep Inception nets, without any hard-coded agreement of dimension, can beat ResNets of similar sizes on ImageNet.
Experiments on ImageNet also show that the proposed grow-push-prune pipeline is able to derive efficient models which can outperform popular fixed nets, other pruned models, small Inception nets during growing, and ResNets at similar complexities.

\section*{Acknowledgments}
This research was enabled in part by support provided by Compute Canada (\url{www.computecanada.ca}), the Natural Sciences and Engineering Research Council of Canada, and McGill Engineering Doctoral Awards.

\bibliographystyle{model2-names}
\bibliography{ycviu-template-with-authorship-referees}

\end{document}